\pdfoutput=1

\documentclass[11pt]{article}
\usepackage{authblk}
\usepackage[]{acl}
\DeclareRobustCommand{\disambiguate}[3]{#2}

\usepackage{times}
\usepackage{latexsym}
\usepackage{stmaryrd}
\usepackage{enumitem}
\usepackage[most]{tcolorbox}
\usepackage{bm}
\usepackage{subfigure}
\usepackage{subcaption}
\usepackage[T1]{fontenc}

\usepackage[utf8]{inputenc}

\usepackage{microtype}
\usepackage{bbm}
\usepackage{graphicx}
\usepackage{hyperref}
\usepackage{csquotes}
\usepackage{xcolor}
\usepackage{adjustbox}
\usepackage{makecell}
\usepackage{multirow}
\usepackage{amsmath}
\usepackage{amssymb}
\usepackage{inconsolata}
\usepackage{pifont} 

\usepackage{algorithm}
\usepackage[noend]{algorithmic}
\usepackage{amsmath}

\title{Paraphrase and Solve: Exploring and Exploiting the Impact of Surface Form on Mathematical Reasoning in Large Language Models}

\author{\textbf{Yue Zhou}$^1$\thanks{\hspace{1mm}This work was done during the first author's internship at MIT-IBM Watson AI Lab.} \quad
        \textbf{Yada Zhu}$^2$ \quad
        \textbf{Diego Antognini}$^2$ \quad
        \textbf{Yoon Kim}$^3$ \quad
        \textbf{Yang Zhang}$^2$\\
       
  $^1$University of Illinois Chicago\quad $^2$MIT-IBM Watson AI Lab, IBM Research\quad $^3$MIT CSAIL\\ 
  {\tt yzhou232@uic.edu, yzhu@us.ibm.com} \\
  {\tt yoonkim@mit.edu, \{diego.antognini, yang.zhang2\}@ibm.com}\\
  
}

\begin{document}
\maketitle

\begin{abstract}

This paper studies the relationship between the surface form of a mathematical problem and its solvability by large language models. We find that subtle alterations in the surface form can significantly impact the answer distribution and the solve rate, exposing the language model's lack of robustness and sensitivity to the surface form in reasoning through complex problems. To improve mathematical reasoning performance, we propose Self-Consistency-over-Paraphrases (SCoP), which diversifies reasoning paths from specific surface forms of the problem. We evaluate our approach on four mathematics reasoning benchmarks over three large language models and show that SCoP improves mathematical reasoning performance over vanilla self-consistency, particularly for problems initially deemed unsolvable. 
Finally, we provide additional experiments and discussion regarding
problem difficulty and surface forms, including cross-model difficulty agreement and paraphrasing transferability, and Variance of Variations (VOV) for language model evaluation. 

\end{abstract}
\section{Introduction}

Despite the impressive performance of  large-scale language models (LLMs) across many tasks, their ability to reason through complex problems such as mathematics remains a bottleneck~\cite{LLM-eval2, LLM-eval1, LLM-eval3}; they can solve problems that are challenging for humans but can also struggle with seemingly simple ones. This raises the following question: what factors contribute to the difficulty of a math problem for an LLM? 

Specifically, the information in a math problem can be divided into two types. The first is the \emph{semantic  information}, which involves the problem specification and knowledge involved in the math problem. The second is the \emph{surface form}, \emph{i.e.}, how the questions, assumptions, and constraints are described in the math problem. Intuitively, the semantic information should be the primary determining factor of the difficulty of the math problem, and surface form should only have a marginal impact. This paper investigates the extent to which this is true for LLMs.

\begin{figure*}[!ht]
    \centering 
    \includegraphics[width=0.95\textwidth]{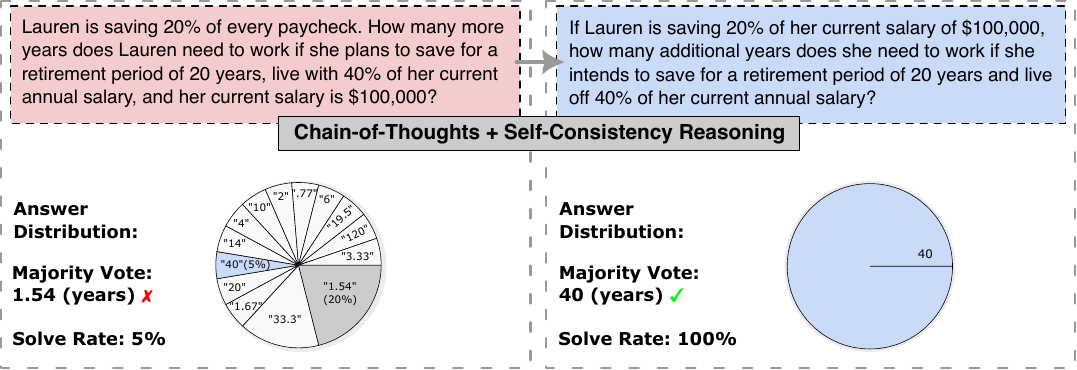}
    \caption{Comparison of the answer distribution and solve rate between surface form variations of a math word problem from GSM8K, when prompted to GPT-3.5-turbo using Self-Consistency, with 40 sampled reasoning paths. Solve rate can vary dramatically between surface forms with equivalent semantics.
}
    \label{fig:ts}
\end{figure*}

In this paper, we explore into the relationship between the problem's surface form and its solvability by an LLM. Specifically, we follow the self-consistency setting~\cite{SC} to sample multiple answers to the same math problem and compute \emph{solve rate} as the percentage of correct answers. Our primary finding is that, counter-intuitively, subtle alterations in the surface form of a math problem can significantly impact the answer distribution and solve rate. 
Consider the example in Figure~\ref{fig:ts}, where the left and right panels contain an identical math problem described in two different ways.  Despite the no change in problem semantics,  the solve rate increases from 5\% to 100\%, with all reasoning paths leading to the correct answer - what initially appears to be a difficult problem to the language model transforms into an easily solvable one. 
This phenomenon exposes the language model's lack of robustness and sensitivity to the surface form in reasoning through complex problems. 

Motivated by this finding, we propose to improve the mathematical reasoning performance of the language model by diversifying reasoning paths from specific surface forms of the problem. We leverage the language model's paraphrasing ability to generate surface forms with identical semantics\footnote{Rigorously, the surface forms can be regarded as ``quasi-paraphrases that convey approximately the same meaning using different words''~\cite{paradefine}.} and propose \textbf{Self-Consistency-over-Paraphrases (SCoP)}, which consists of two steps: \ding{182} For each math problem, generate $K$ paraphrase using an LLM; and \ding{183} Ask the LLM to generate $N/K$ reasoning paths for each paraphrase, and then select the most consistent answer among the $N$ answers.
The intuition is that if a problem exhibits a low solve rate and ineffective reasoning paths due to its original surface form, introducing diversity in its surface forms can be beneficial. We also introduced in-context exemplars to the language model when paraphrasing, which are the paraphrases that obtain a solve rate improvement over their original problem, aiming to generate surface forms with the same semantics yet a higher solve rate through language models' in-context learning abilities~\cite{ice1,ice2}. 

We evaluate our approach on four mathematics reasoning benchmarks: GSM8K~\cite{GSM8K}, AQuA~\cite{aqua}, MATH~\cite{math}, and MMLU-Math~\cite{mmlu}, over three large language models: LLaMA-2-70b~\cite{llama2}, GPT-3.5-turbo and GPT-4~\cite{openai}. Our experiments show that SCoP improves mathematical reasoning performance over vanilla Self-Consistency, particularly for problems initially deemed unsolvable. In additional experiments, we show that the difficulty ranks across language models are positively correlated, with higher agreement within the GPT model family and simpler datasets. We propose Variance of Variations (VOV) as a metric for evaluating language model robustness against surface form variations. Finally, we explain why SCoP can be effective using a data difficulty map based on the entropy of answer distribution and the solve rate. Our code is publicly available.\footnote{ \url{https://github.com/Yue-LLM-Pit/SCoP/}}


\section{Problem Difficulty and Surface Forms} 
\label{sec:pilot}

In this section, we present our pilot study of the impact of surface form on LLMs' ability to solve the problem. In all our studies, we follow the self-consistency setting~\cite{SC}, which extends over chain-of-thought~\cite{cot} by using sampling to generate a variety of reasoning paths. From this setting, we quantify the \emph{difficulty} of a problem \emph{w.r.t} a language model by its \textbf{solve rate}, which is the proportion of the reasoning paths that lead to the correct answer.
 
When the solve rate exceeds 50\%, a majority vote guarantees the correct answer. Note the solve rate measures the difficulty of a single problem input and is also a model-dependent metric. 

To study how surface form impacts the solve rate, we use the math word problem from the GSM8K dataset~\cite{GSM8K}. For each math problem, we generate a paraphrase using GPT-3.5-turbo\footnote{\url{https://platform.openai.com/docs/models/gpt-3-5}} (detailed instructions are shown in Appendix~\ref{sec:temp}). We then compare the solve rates of the original problem and the paraphrase solved by GPT-3.5-turbo using self-consistency with $N = 40$ and a temperature of $0.7$.

Our finding is that the solve rate varies significantly across the surface forms. Figure~\ref{fig:ts} shows an example with the original problem on the left and the paraphrased one on the right. In the original problem, the reasoning paths result in a disarrayed answer distribution, with merely 5\% achieving the correct answer ``40'' and the aggregated answer ``1.54'' (20\%). In contrast, the solve rate of the paraphrase problem is 100\%. We have identified many more such examples with drastic improvement in solve rate, presented in Table~\ref{tab:qa1}.

\begin{figure}[!t]
    \centering 
    \includegraphics[width=0.94\columnwidth]{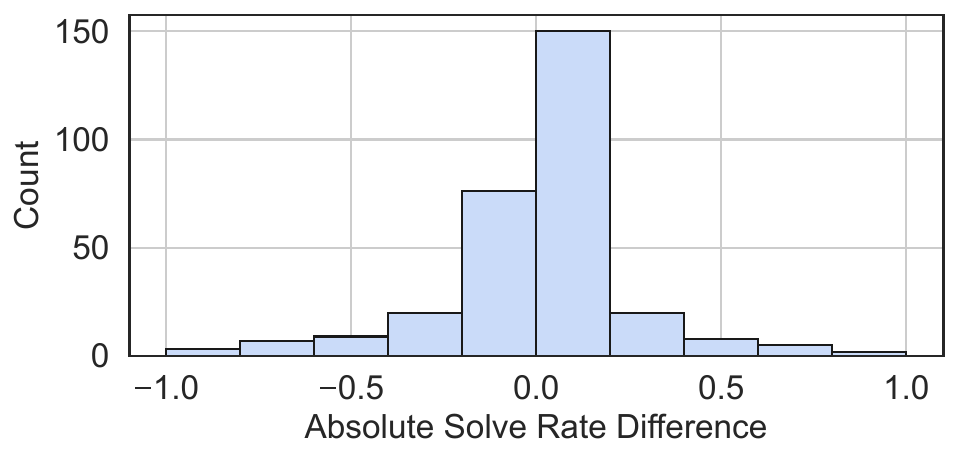}
    \caption{GSM8K - solve rate difference - from original to one of the random naive paraphrases.}
    \label{fig:naive_dist_1}
\end{figure}

We further calculate the histogram of the solve rate changes in the paraphrased problem compared to the original one, shown in Figure~\ref{fig:naive_dist_1}. As can be observed, the distribution is heavy-tailed, with 11.7\% of the paraphrases resulting in over 25\% absolute improvement in solve rate and with 13\% resulting in over 25\% absolute deterioration. 

This phenomenon exposes the language model's lack of robustness and sensitivity to a comprehensive problem's surface form. It suggests that the challenge of some problems may not be due to the model's limitations, but rather the ineffective generation of reasoning paths from certain surface forms. We therefore seek to take advantage of this phenomenon to improve language model reasoning through surface form modifications, mirroring the way paraphrasing aids a student's cognitive and problem-solving processes~\cite{shengun}.

\section{Self-Consistency over Paraphrases}

Motivated by the findings in Section~\ref{sec:pilot}, we propose a framework called \textbf{Self-Consistency-over-Paraphrases (SCoP)}, which leverages the LLMs to generate paraphrases of math problems to improve their ability in solving them.

\subsection{Framework Overview} 

\begin{figure}[!t]
    \centering 
    \includegraphics[width=0.94\columnwidth]{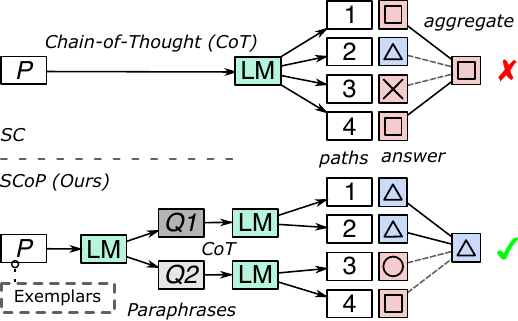}
    \caption{A comparison between Self-Consistency and our SCoP. SCoP splits $N$ reasoning paths over $K$ in-context learned paraphrases, instead of devoting all $N$ reasoning paths to the single original problem $P$. The final answer is selected by aggregating all reasoning paths from these paraphrases with a majority vote.}
    \label{fig:ts2}
\end{figure}

As shown in Figure~\ref{fig:ts2}, SCoP consists of two steps.

\noindent \textbf{$\bullet$ Step 1: Paraphrase.} Prompt the LLM to generate $K$ paraphrases of the original problem. For notational ease, denote $p$ as the original problem, and $\bigcup_{k=1}^K \{q_k\}$ as the $K$ paraphrases.

\noindent \textbf{$\bullet$ Step 2: Solve.} 
For each paraphrase, we ask the LLM to generate $N/K$ reasoning paths, and thus the total number of generated answers is $N$. 
We then select the most consistent answer across the $N$ reasoning paths as the final answer.

The intuition behind SCoP is that if a problem exhibits a low solve rate and ineffective reasoning paths due to its original surface form, introducing diversity in its surface forms would be beneficial.

There are two important notes regarding SCoP. First, when we increase $K$, the total number of reasoning paths $N$ is held fixed, which separates the effect of increasing the diversity of reasoning paths from increasing the number of reasoning paths. This also ensures a fair comparison with other self-consistency baselines.

Second, there are two procedures in SCoP that involve an LLM, one to generate paraphrases (Step 1) and one to generate answers (Step 2). We use the \emph{same} LLM to perform both tasks. In this way, we can ensure that any performance improvement of SCoP is due to the diversity of paraphrasing itself, rather than cross-sharing of knowledge across different LLMs. In addition, there is no human annotation, training, fine-tuning, or auxiliary models involved in our SCoP framework. 

\subsection{Paraphrase Generation} 

The paraphrase generation in Step 1 is crucial to the success of SCoP. In this work, we explore two paraphrase generation methods.

\vspace{-0.05in}
\paragraph{Na\"ive.} The na\"ive approach instructs the language model to generate $K$ paraphrases of the math problem. However, this could generate many paraphrases with \emph{worse} solve rate, because the solve rate change has high variability in both directions (as shown in Figure~\ref{fig:naive_dist_1}).

\vspace{-0.05in}
\paragraph{In-Context Learning.} To increase the chance of generating `good' paraphrases, we propose an in-context learning approach,\footnote{An alternative can be automatic prompt engineering, see Appendix~\ref{sec:ape}.} where we obtain $N_{shot}$ `good' paraphrases as the in-context exemplars (marked as [\texttt{Exemplars}] in Figure~\ref{fig:ts2}). The `good' paraphrases are formally defined as paraphrases that contribute to a solve rate improvement (by a preset margin $\delta$) over the original problem. To obtain the `good' paraphrases, we first generate some candidate paraphrases using the aforementioned na\"ive approach on a small number of math problems with labeled answers. We then compute the solve rate of the original problem and the paraphrases and select those whose improvement is over the margin $\delta$. The detailed algorithm is presented in Algorithm~\ref{alg:search}.

\begin{algorithm}[!t]
\small
\caption{Paraphrase Exemplar Search}
\label{alg:search}
\begin{algorithmic}[1]
\STATE \textbf{Input:} Training data $D^{tr}$, $N_{shot}$, margin $\delta$. Init. Candidates list $C$.
\FOR{step $t$ in $\{1, 2, \dots, T\}$}

\IF{$\operatorname{\mathrm{Length}}($C$) = N_{shot}$ }
       \STATE \textbf{break}
\ENDIF
       
\STATE 
       
       Sample a problem $p$ from $D^{tr}$ without replacement.
\STATE
       Compute solve rate $\operatorname{\mathrm{SR}}(p)$
       
\STATE
       Obtain $K$ Paraphrases $\{q_1, \dots, q_K\}$ of $p$. 

\FOR{$k=1$ to $K$}
\STATE
       Compute solve rate $\operatorname{\mathrm{SR}}(q_k)$
       \IF{$\operatorname{\mathrm{SR}}(q_k) >= \operatorname{\mathrm{SR}}(p) + \delta$}
       \STATE Add \{$p$, $q_k$\} to Candidates list $C$.
       \STATE \textbf{break}
       \ENDIF

\ENDFOR
\ENDFOR

\end{algorithmic}
\end{algorithm}

\section{Experiments}

\begin{table*}[t!]
\centering
\begin{adjustbox}{width=\textwidth}


\begin{tabular}{llcccccccc}
\hline\hline
     &      & \multicolumn{4}{c}{GPT-3.5-Turbo}                       & \multicolumn{4}{c}{LLaMA-2-70b}                           \\ \cline{2-10}
     &                     & GSM8K        & AQuA         & MATH        & MMLU        & GSM8K        & AQuA         & MATH         & MMLU         \\ \cline{2-10}
 & HPR (\%) & 31.3        & 42.5        & 68.0          & 64.0          & 52.0           & 76.3         & 98.2         & 81.6         \\ \hline
SC &                    & 76.3 (24.5) & 66.9 (22.2) & 59.0 (39.7) & 52.8 (26.3) & 58.7 (20.5) & 40.5 (22.0) & 10.5 (8.9)  & 32.8 (17.4) \\ \hline
 & $k$ = 1              & 72.2 (27.7) & 63.4 (28.9) & 55.0 (37.5) & 48.4 (27.5)  & 51.0 (28.2)    & 38.1 (26.8) & 24.6 (23.2) & 27.2 (20.2) \\ 
SCoP & $k$ = 2          & 76.0 (34.0)  & 65.8 (28.9) & 56.5 (39.7) & 52.8 (32.5)  & 54.3 (26.9) & 39.5 (25.0)    & 29.8 (28.6) & 29.6 (22.3) \\ 
(Na\"ive) & $k$ = 4     & 77.7 (36.2) & 67.3 (29.8) & 57.5 (39.0) & 56.0 (36.3) & 55.7 (32.1) & 41.4 (25.6)  & \textbf{31.6 (30.4)} & 32.0 (24.9) \\ 
 & $k$ = 8              & 79.3 (39.4) & 68.1 (33.5) & 59.5 (43.4) & 55.6 (33.8) & 60.3 (33.3) & 41.4 (25.6)  & 28.1 (26.8) & 35.6 (28.0) \\ \hline
 & $k$ = 1              & 77.9 (39.0) & 66.4 (29.8) & 54.0 (36.8) & 52.5 (32.6) &     58.7 (39.9)         &     42.9 (29.8)         &      23.4 (22.0)        &       34.6 (23.7)     \\ 
SCoP & $k$ = 2 & \textbf{80.5 (39.2)} & 68.5 (31.7) & 57.5 (39.1) & 55.5 (34.1) &    59.3 (36.3)          &         43.7 (30.4)     &     24.6 (23.2)        &      37.6 (26.3)      \\
(ICL$_\text{para}$)      & $k$ = 4 & 79.2 (38.3) & \textbf{70.5 (35.4)} & 58.0 (41.2) & \textbf{58.0 (39.5)}   &     61.7 (40.5)         &       44.5 (30.4)       &     26.9 (25.6)         &      \textbf{37.8 (26.5) }     \\
 & $k$ = 8              & 80.2 \textbf{(40.6)} & 69.7 (34.4) & \textbf{60.0 (44.1)} & 56.5 (34.9) &     \textbf{ 63.3 (40.5)}        &     \textbf{ 46.5 (31.9) }       &    25.2 (23.8)          &     37.6 (25.8)         \\ \hline
\end{tabular}

\end{adjustbox}
\caption{Accuracy of SCoP distributing $N/K$ reasoning paths over $K$ in \{1, 2, 4, 8\} paraphrases in Na\"ive and ICL$_\text{para}$ settings, against Self-Consistency (SC). Hard Problem Ratio (HPR\%) represents the percentage of problems with an original solve rate $\leq 0.5$ by Self-Consistency (SC). Accuracy is reported for both Hard Problems (HPR\% $\leq 0.5$) (inside parentheses) and global accuracy across the entire dataset (outside parentheses).} 
\label{tab:res1}
\end{table*}

\begin{table*}[ht!]
\centering
\begin{adjustbox}{width=0.75\textwidth}

\begin{tabular}{clcccc}
\hline\hline                                       
  Model   &     & GSM8K        & AQuA         & MATH        & MMLU               \\ \hline

& HPR (\%) &   56     &    75.2     &    95.2      &       81.6           \\ \cline{2-6}
LLaMA-2-70b & Self-Consistency  &  61.1 (30.0)  &  44.1 (25.7)  &  13.4 (9.2)  &  34.4 (19.6)  \\ 
 & SCoP (ICL$_\text{para}$, $k$ = 8) &  \textbf{65.1 (38.6)}   &   \textbf{48.8 (33.5)}  &  \textbf{23.6 (20.2)}   &     \textbf{36.4 (27.0)}  \\ \hline 

  & HPR (\%) & 22        & 36      & 75          & 62                 \\ \cline{2-6}
GPT-3.5-Turbo & Self-Consistency         & 80.0 (9.1) & 70.0 (16.7) & 51.6 (29.8) & 54.4 (26.9)  \\ 
 & SCoP (ICL$_\text{para}$, $k$ = 8)  & \textbf{82.0 (36.4)} & \textbf{74.0 (27.8)} & \textbf{57.6 (38.2)} & \textbf{58.4 (36.5)}     \\ \hline
 
 & HPR (\%) &    4    &    18     &     58     &         38         \\ \cline{2-6}
GPT-4 & Self-Consistency  &  98.0 (50.0)  &  84.0 (11.1)  &  64.0 (37.9)  &  74.0 (31.6)  \\ 
 & SCoP (ICL$_\text{para}$, $k$ = 8) &  98.0 (50.0)   &   \textbf{86.0 (33.3)}  &  \textbf{66.0 (41.4)}   &   \textbf{78.0 (57.9)}  \\ \hline 
\end{tabular}

\end{adjustbox}
\caption{A comparison of the performance (accuracy) between SC and SCoP (ICL$_\text{para}$ paraphrasing, with $k = 8$) using 4-shot in-context chain-of-thought exemplars over three language models. Accuracy is reported for both Hard Problems (HPR\% $\leq 0.5$) (inside parentheses) and global accuracy across the entire dataset (outside parentheses).} 
\label{tab:res2}
\end{table*}

In this section, we will describe our experiment results evaluating the effectiveness of SCoP, as well as additional studies on how SCoP works. 

\subsection{Experimental Settings}

\paragraph{Datasets} We evaluate our approach on the following public mathematics reasoning benchmarks:

\noindent \textbf{$\bullet$ GSM8K}~\cite{GSM8K} contains 8.5K linguistically diverse grade school-level math questions with moderate difficulties. 

\noindent \textbf{$\bullet$ AQuA}~\cite{aqua} consists of 100K algebraic word problems, including the questions, the possible multiple-choice options, and natural language answer rationales from GMAT and GRE.

\noindent \textbf{$\bullet$ MATH}~\cite{math} is a competition mathematics dataset containing 12,500 problems with challenging concepts such as Calculus, Linear Algebra, Statistics, and Number Theory.

\noindent \textbf{$\bullet$ MMLU}~\cite{mmlu} is a comprehensive dataset containing various subjects. We specifically utilized the mathematics section of the dataset, which comprises college and high-school-level mathematics, statistics, and abstract algebra. 

\paragraph{Language Models} We utilize three popular LLMs trained with RLHF~\cite{RLHF}: LLaMA-2 (70B)~\cite{llama2}, an open-source LLM by Meta AI, GPT-3.5-turbo (version 0613), and GPT-4~\cite{openai}, accessed via the OpenAI API. All experiments are conducted in zero-shot or few-shot settings, without training or fine-tuning the language models. We choose the temperature $T = 0.7$ and Top-p $ = 1.0$ for sampling decoding for all three language models. The total number of reasoning paths $N$ we sample for each problem is $40$, following~\citet{SC}. 

\paragraph{Implementation Details} For paraphrase generation (Step 1), we evaluate the two aforementioned schemes \ding{182} \textbf{Na\"ive}: We use the template `` \emph{Paraphrase the following math problem:} \texttt{\{question\}}'' to prompt the language model to paraphrase the original problem; \ding{183} \textbf{In-Context Learning ($\text{ICL}_\text{para}$)}: We randomly select a set of 8 paraphrase exemplars by Algorithm~\ref{alg:search} with margin\footnote{We performed an ablation study of the margin effect on a separate development sets and found that using an extremely large margin can damage performance. See Appendix~\ref{sec:app-margin}.} $\delta=0.3$. The details of the prompt templates are available in Appendix~\ref{sec:temp}.

For answer generation (Step 2), we also implement two schemes: \ding{182} \textbf{Zero-Shot Chain-of-Thought (CoT)}~\cite{cot0}, which appends ``\emph{Let's think step by step.}'' to the question text; and \ding{183} \textbf{Four-Shot CoT}, where we append four-shot in-context examples with CoT to the LLM when \emph{solving} the math problems. Note that the in-context examples for answer generation are different in functionality and format from the ones for $\text{ICL}_\text{para}$. 

\subsection{Main Results} 

\paragraph{Zero-Shot CoT} Table~\ref{tab:res1} illustrates the performance of SCoP under the zero-shot CoT setting, compared with the vanilla self-consistency (SC), using LLaMA-2-70b and GPT-3.5-turbo. We vary the number of paraphrases $K$ across \{1, 2, 4, 8\} while keeping the total number of reasoning paths fixed as 40. Due to resource constraints, we randomly sampled 300 data points from each test set, except for AQuA, which contains 254 testing examples.

The performance metric is the accuracy of the self-consistency answer. We also report the accuracy over hard problems, defined as the problems whose original accuracy is below 50\%. The accuracies over all problems and hard problems are reported inside and outside parentheses respectively. HPR\% (Hard Problem Ratio) denotes the percentage of such hard problems.


There are three general observations. First, SCoP with the two paraphrasing schemes both outperform the vanilla self-consistency baseline. Surprisingly, even the na\"ive paraphrasing can lead to performance improvement, despite the high chances of generating paraphrases with a worse solve rate (see Figure~\ref{fig:naive_dist_1}). We will discuss a hypothesis in Section~\ref{sec:discuss}. Between the two schemes, ICL$_\text{para}$ consistently outperforms Na\"ive. Second, the performance improvement generally increases as $K$ increases. Third, more significant performance gain over LLaMA-2-70B.


The results further indicate that MATH and MMLU are considerably more challenging than GSM8K and AQuA, as evidenced by their high HPR\% and low overall accuracy. Moreover,  significant accuracy gains are from the original ``Hard Problems'', suggesting that changing surface forms can solve the problems initially deemed unsolvable by self-consistency. Finally, when solving the MATH dataset with LLaMA-2-70b, ICL$_\text{para}$ underperforms Naïve paraphrasing. We hypothesize that the MATH problems present a significant challenge for LLaMA-2-70b, making it difficult to effectively learn paraphrasing from in-context examples.

\paragraph{Four-Shot CoT}

One caveat of the zero-shot CoT results is that SCoP (ICL$_\text{para}$) has indirect access to additional ground-truth information from in-context exemplars. There is also a question of whether the advantage of SCoP over SC will diminish as both are exposed to more examples. To ensure a fair comparison and further validate the effectiveness of SCoP,  Table~\ref{tab:res2} shows results under the four-shot CoT setting, where the baselines also have access to some ground-truth answer information. Due to resource constraints, we evaluate GPT4 with 100 random samples from each dataset. The results show that while four-shot CoT can improve SC and SCoP in general (compared with zero-shot CoT), SCoP still consistently outperforms SC over all three language models. 
The only exception is GPT4 on GSM8K, which already achieves near-perfect performance with SC, thus SCoP only achieves equivalent performance.

\subsection{Additional Studies}

\begin{figure*}
    \centering
    \subfigure[]{\includegraphics[width=0.24\textwidth]{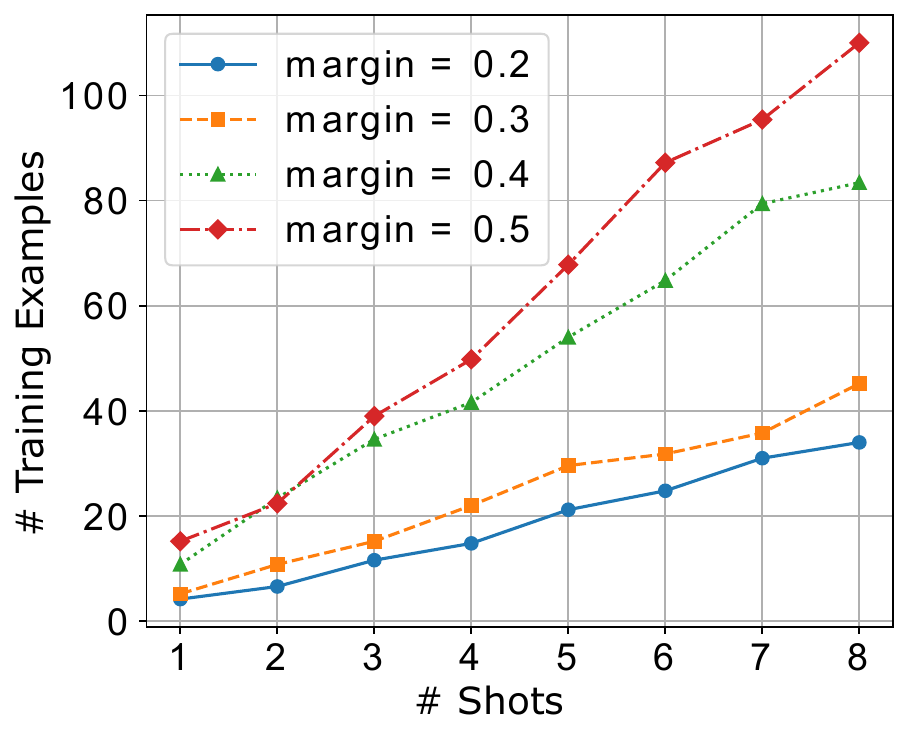}}
    \subfigure[]{\includegraphics[width=0.24\textwidth]{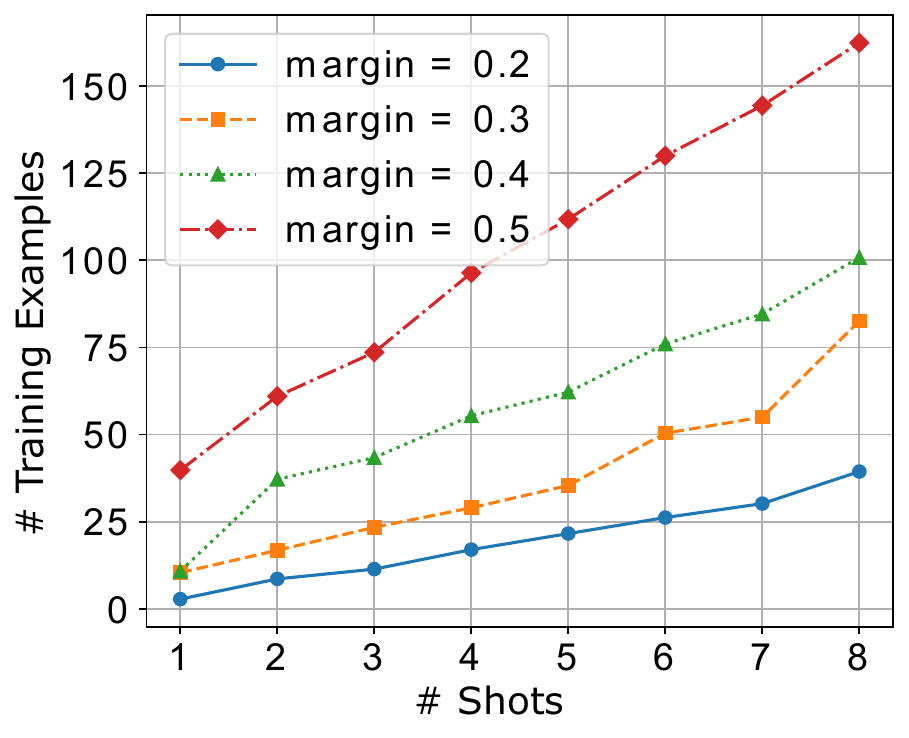}} 
    \subfigure[]{\includegraphics[width=0.24\textwidth]{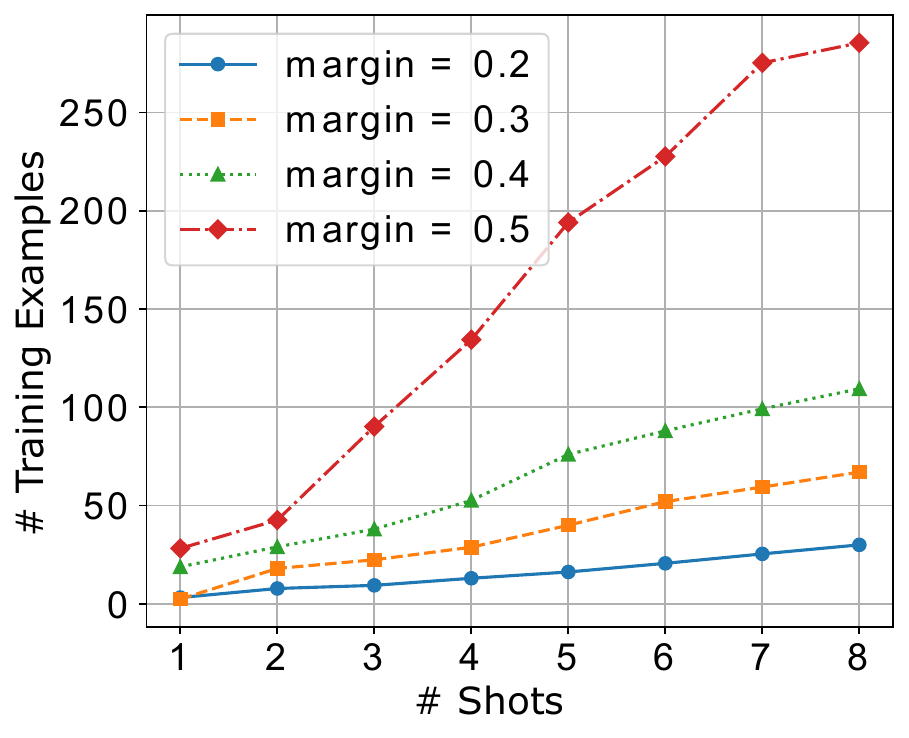}}
    \subfigure[]{\includegraphics[width=0.24\textwidth]{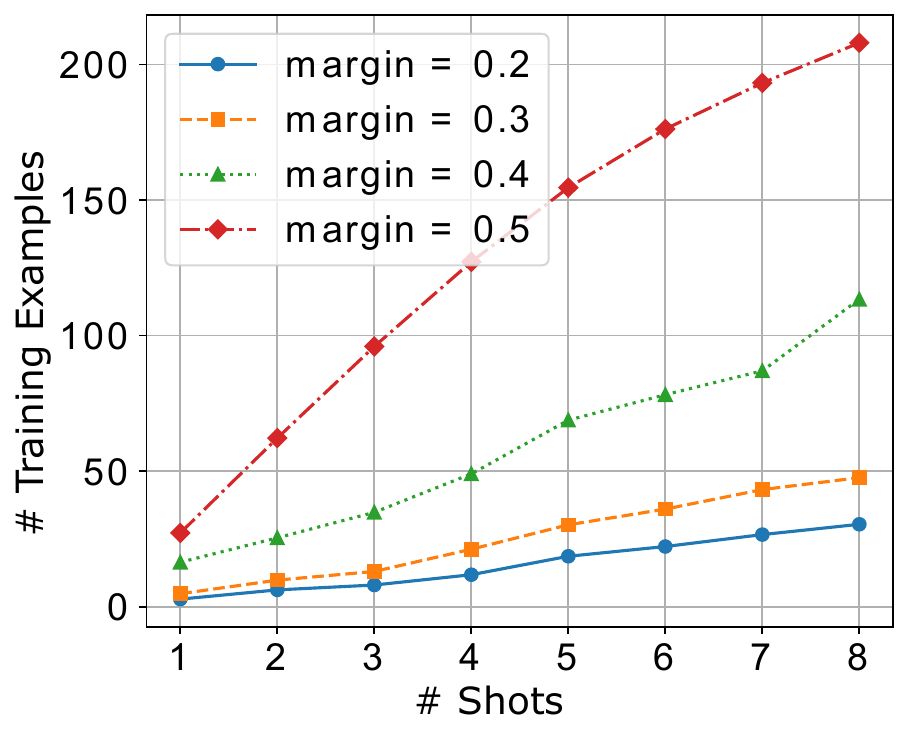}}
    \caption{(a) GSM8K (b) AQuA (c) MATH (d) MMLU. The average number of data points in the training set needed for obtaining $N_{shot}$ exemplars at different margins.}
    \label{fig:find_ices}
\end{figure*}

\paragraph{Searching for Exemplars} 
Since our in-context learning paraphrasing scheme requires access to ground-truth answers, we would like to study how many problems with ground-truth answers are needed. Figure~\ref{fig:find_ices} illustrates how many data points in the training set, on average, need to be sampled to obtain $N_{shot}$ `good' paraphrases (x-axis) with different margins.
We can observe that, although satisfying a large margin requires more samples, it is relatively easy (typically every $\pm 5$ example) to find a sample that substantially improves the solve rate after paraphrasing. This, again, indicates the sensitivity of the language model to surface form variations in mathematical reasoning.

\paragraph{Difficulty Beliefs Across Language Models} 
An intriguing question is how different language models rank the difficulty of the problems. We measure the agreement between language models on problem difficulty by Spearman's rank correlation
of the solve rate for original problems across four datasets. As shown in Table~\ref{tab:x1}, the ranks of the difficulty (by solve rate) are all positively correlated. However, the degree of correlation varies, with higher agreement observed within the GPT model family and on simpler datasets. 

\begin{table}[!t]
\centering
\begin{adjustbox}{width=\columnwidth}

\begin{tabular}{llll}
\hline\hline
      & GPT3.5, GPT4 & GPT3.5, LLaMA-2 & GPT4, LLaMA-2 \\ \hline
GSM8K & 0.573**        & 0.649***          & 0.445*          \\
AQUA  & 0.543***       & 0.227***          & 0.314*          \\
MATH  & 0.554***       & 0.242*            & 0.433*          \\
MMLU  & 0.313*         & 0.320***          & 0.233          \\ \hline
\end{tabular}

\end{adjustbox}
\caption{Spearman's rank correlation of original problems' solve rate across language models.} 
\label{tab:x1}
\end{table}

\paragraph{Paraphrase Transfer}

We investigate whether paraphrases from a stronger LLM can be transferred to weaker ones and improve SCoP. Table~\ref{tab:cross} demonstrates the paraphrase transfer performance of SCoP (Na\"ive, $k$ = 8) on 100 randomly sampled data points from MMLU and GSM8K under the zero-shot CoT setting. In general, paraphrases produced by GPT-4 can be utilized by GPT3.5-turbo or LLaMA-2-70b for further performance improvements, with an exception with LLaMA-2 on MMLU, where GPT4 and LLaMA-2 exhibit the lowest Spearman rank correlation of solve rate. We hypothesize that the benefits of transferring paraphrases across models may depend on the agreement in their beliefs of problem difficulty.


\begin{table}[t!]
\centering
\begin{adjustbox}{width=0.8\columnwidth}

\begin{tabular}{llll}
\hline\hline
Solver & Paraphraser & MMLU  & GSM8K \\ \hline
GPT-3.5  & Self & 50.0     & 78.0      \\
GPT-3.5  & GPT-4 & \textbf{54.0}     & \textbf{84.0}      \\ \hline
LLaMA-2  & Self & \textbf{37.0}     &   61.0   \\
LLaMA-2  & GPT-4 & 34.0     &  \textbf{69.0}  \\ \hline
\end{tabular}

\end{adjustbox}
\caption{Performance of SCoP (Na\"ive, $k$ = 8) on MMLU and GSM8K, with different paraphrasers.} 
\label{tab:cross}
\end{table}

\begin{figure*}[!ht]
    \centering
    \subfigure[]{\includegraphics[width=0.325\textwidth]{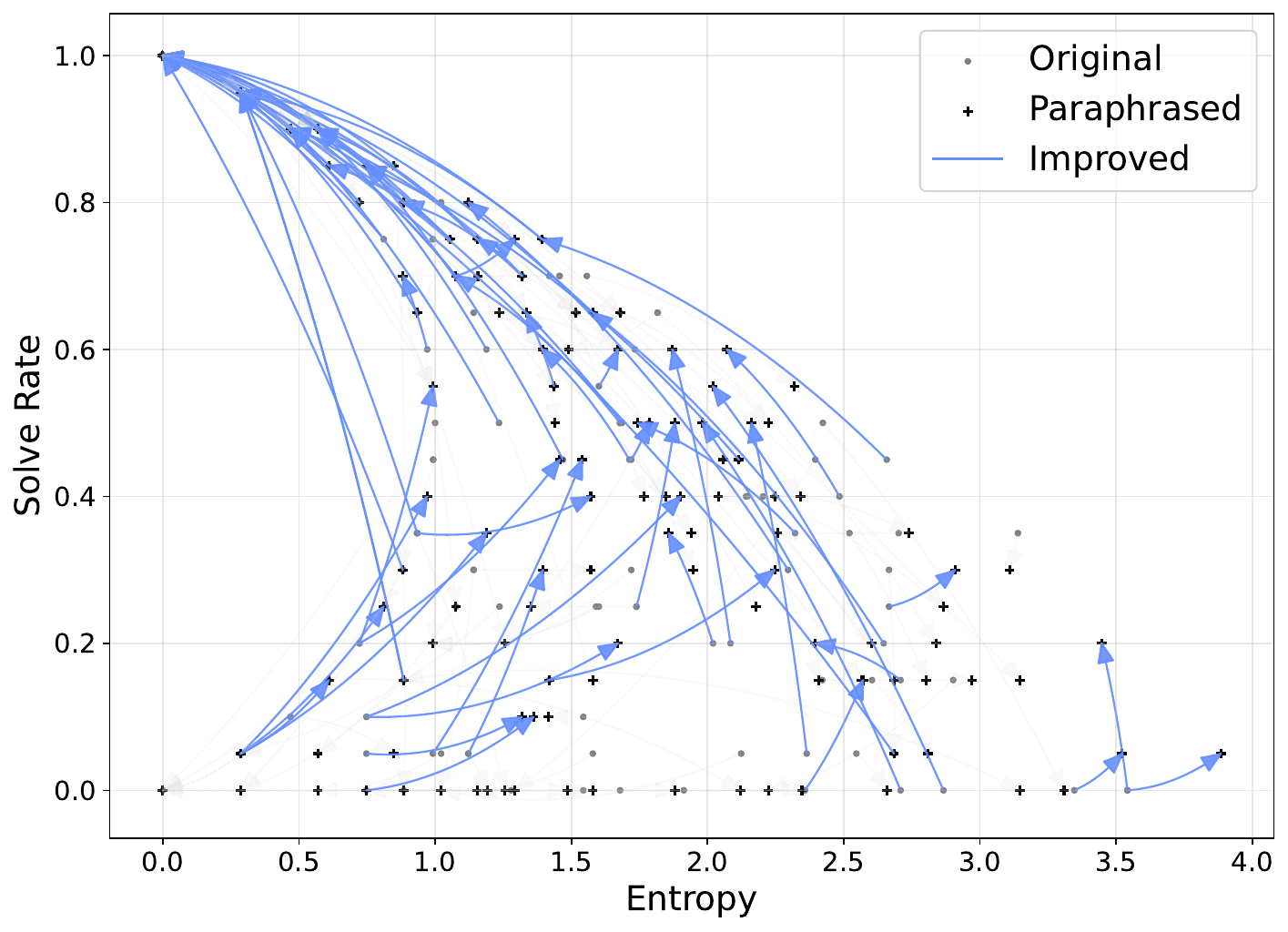}}
    \subfigure[]{\includegraphics[width=0.325\textwidth]{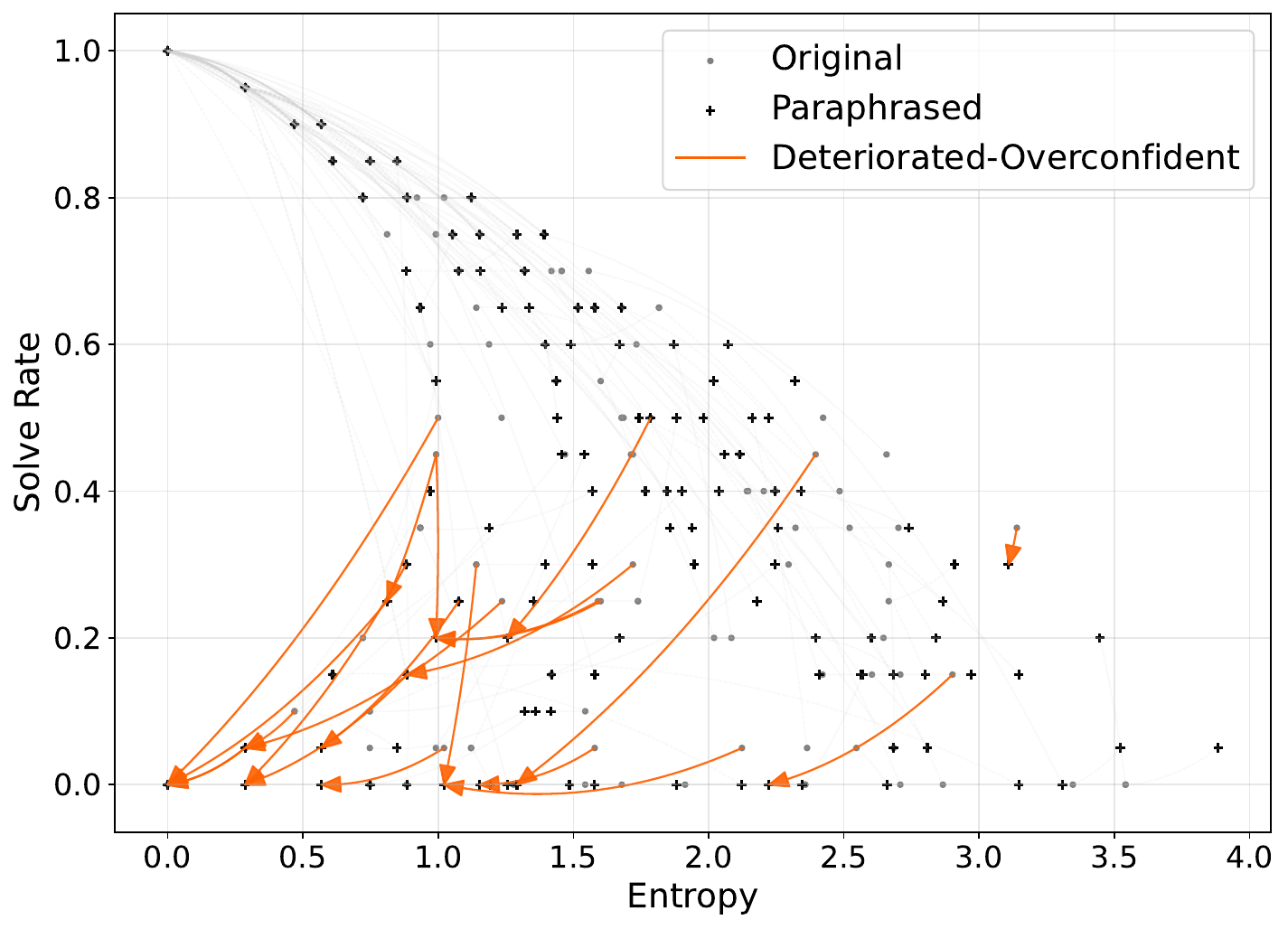}} 
    \subfigure[]{\includegraphics[width=0.325\textwidth]{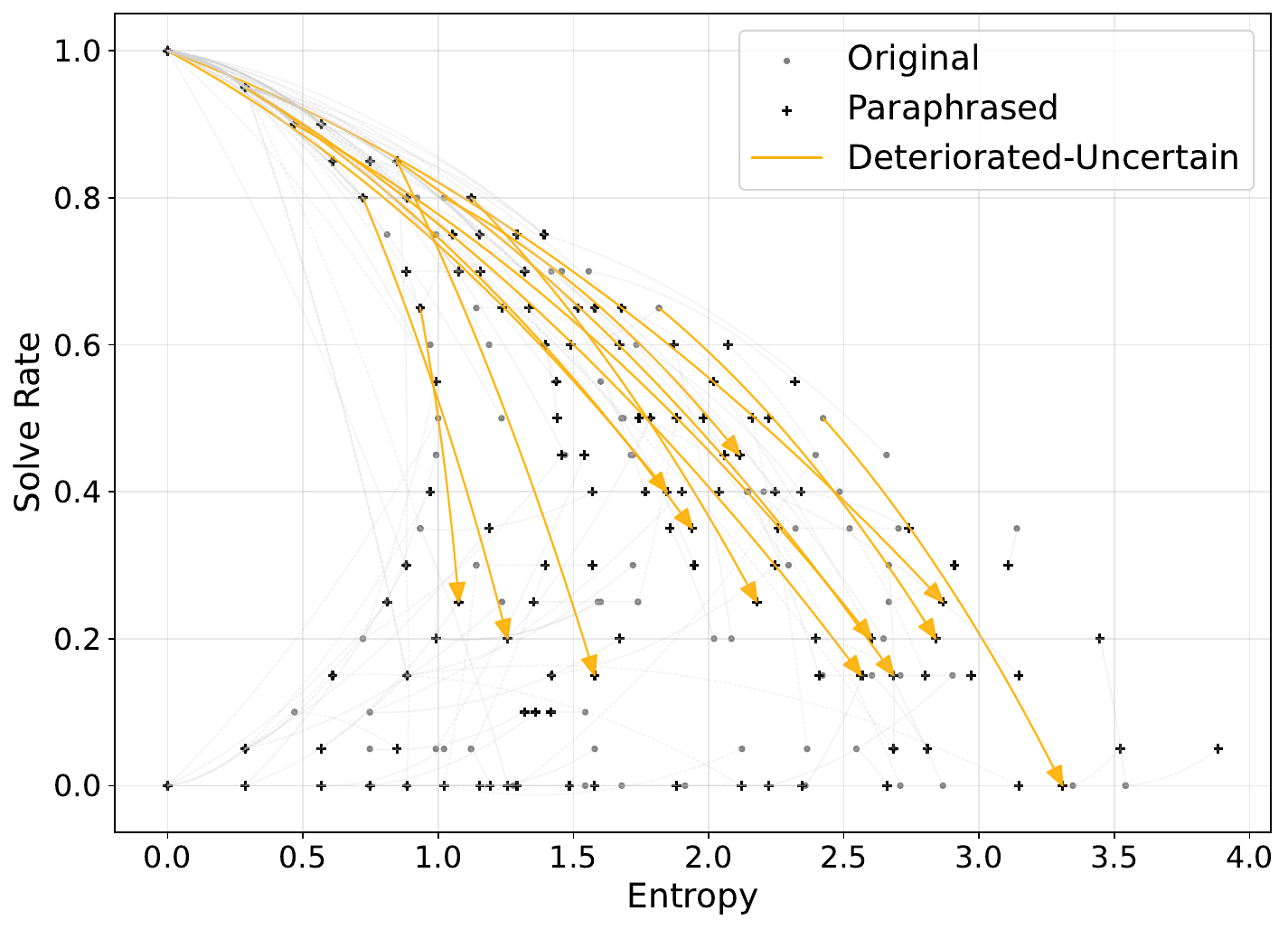}}
    \caption{Data Difficulty Map for GSM8K using GPT3.5, with three types of changes from solving the original problem to one of its random paraphrases: (a) Improvement, (b) Overconfidence, and (c) Uncertainty. Arrows indicate the solve rate and entropy change from solving the original problem to its paraphrased version.}
    \label{fig:dis1}
\end{figure*}

\paragraph{Variance of Variations} 

In light of the considerable variability observed in solve rates among problem surface forms (Figure~\ref{fig:naive_dist_1}), we propose and advocate \textbf{Variance of Variations (VOV)} for evaluating language models on reasoning robustness. Let $X(p) \in [0, 1]$ be the random variable representing the solve rates of various paraphrases of a problem $p$. Then the VOV value of the dataset $D$ is then defined as:
\begin{equation}
\operatorname{\mathrm{VOV}}=\mathbb{E}_{p \sim D}[\mathrm{Var}(X(p))]
\label{eq:vov}
\end{equation}

\noindent where $\mathrm{Var}(\cdot)$ is the variance. A large value of VOV indicates high variability in the language model's reasoning ability against problem surface forms. We compute VOV using the solve rate for the $k = 8$ paraphrases and the original problem as $X(p)$ for each $p$. As shown in Table~\ref{tab:vov}, while VOV decreases when a robust model solves a more manageable dataset (\emph{e.g.}, GPT-4 on GSM8K), and ICL$_\text{para}$ generated paraphrases can generally reduce VOV, VOV remains unreasonably high over more challenging datasets and all language models. 

\begin{table}[t!]
\centering
\begin{adjustbox}{width=0.9\columnwidth}

\begin{tabular}{llllll}
\hline \hline
                        &  & GSM8K & AQuA  & MATH  & MMLU  \\ \hline
\multirow{2}{*}{LLaMA2} & Na\"ive & 20.3 & 17.5 & 12.9 & 17.5 \\
                        & ICL$_\text{para}$ & 18.9 & 15.7  &   12.2    & 16.6 \\ \hline
\multirow{2}{*}{GPT-3.5} & Na\"ive & 20.6 & 16.1 & 15.8 & 16.9 \\
                        & ICL$_\text{para}$ & 16.2 & 10.7 & 15.6 & 15.6 \\ \hline
GPT-4                    & ICL$_\text{para}$ & 9.7  & 11.5 & 17.0 & 21.3 \\ \hline
\end{tabular}

\end{adjustbox}
\caption{VOV values across datasets and language models, shown as standard deviation.} 
\label{tab:vov}
\end{table}

\paragraph{Examples of `Good' Paraphrases}
We provide some qualitative examples comparing the solve rates between the original problem and a paraphrased version in Table~\ref{tab:qa1}. It is difficult to visually tell what contributes to a good paraphrase. We will publish these data to encourage future research.

\section{Discussion}
\label{sec:discuss}

We have an intriguing observation that even the na\"ive scheme of generating math paraphrases can improve the overall accuracy. However, the na\"ive scheme has a significant chance of generating worse paraphrases. Why would aggregating over the mixture of better and worse paraphrases still significantly improve the performance?

To explain this, Figure~\ref{fig:dis1} shows three scatter plots of the solve rate against the entropy of answer distributions. The outcome of solving each random paraphrase is represented as a black dot. As can be observed, the dots roughly form a triangular region. The top left corner represents the ideal case with high solve rates and high confidence. The bottom corners, on the other hand, represent two failure modes. The bottom right corner represents the case with low solve rates and low confidence, and the bottom left corner with low soft rates but high confidence (commonly known as over-confidence).

The blue arrows in Figure~\ref{fig:dis1}(a) visualize the cases where the paraphrases improve the solve rate, and they mostly point to the top-left corner. The arrows in Figures~\ref{fig:dis1}(b) and (c) represent the cases where the paraphrases lower the solve rate, and we can observe that the arrows pointing to the bottom right corner (yellow arrows in (b)) far outnumber those to the bottom left corner (red arrows in (c)). This indicates that while the `good' paraphrases would sharpen the answer distribution, the `bad' paraphrases mostly would flatten the distribution. Since the final aggregated answer distribution is predominantly influenced by the sharp distributions, the damage brought by the ``bad'' paraphrases is small compared to the benefit brought by the `good' paraphrases, and thus the aggregate effect across all the paraphrases is still positive.


\begin{table*}[ht!]
\begin{adjustbox}{width=\textwidth}

\begin{tabular}{lllll}
\hline
\textbf{Problem} & \textbf{Source}    & \textbf{Label}    & \textbf{SR}  & \textbf{Voted (F.)} \\ \hline
\makecell[cl]{\textbf{Original}: Jenna has 4 roommates. Each month the electricity bill is \$100. How much will each \\ roommate pay per year for electricity, if they divide the share equally?} & \multirow{3}{*}{GSM8K} & \multirow{3}{*}{"240"} & $0.15$ & \textcolor{red}{"300" ($0.8$)}     \\  
\makecell[cl]{\textbf{Paraphrased}: Jenna shares an apartment with 4 other people. The electricity bill is \$100 per month. \\ If they split the bill equally, how much will each roommate contribute towards the electricity bill in a year?} &  &  & $0.95$ & \textcolor{blue}{"240"}      \\ \hline

\makecell[cl]{\textbf{Original}: Jenny goes to the florist to buy some flowers. Roses cost \$2 each and \$15 for a dozen. \\ If she bought 15 roses and arrived with five 5 dollar bills and they only have quarters for change, \\ how many quarters does she leave with? } & \multirow{4}{*}{GSM8K} & \multirow{4}{*}{"16"} & $0.0$ & \textcolor{red}{"20" ($0.3$)}     \\  
\makecell[cl]{\textbf{Paraphrased}: Jenny visits the flower shop to purchase flowers. She can buy roses individually for \$2 each \\ or buy a dozen roses for \$15. Jenny decides to buy 15 roses in total. She pays with five \$5 bills and \\ the florist can only give her change in quarters. The question asks how many quarters Jenny receives as change.} &  &  & $0.55$ & \textcolor{blue}{"16"}      \\ \hline

\makecell[cl]{\textbf{Original}: Assistants are needed to prepare for preparation. Each helper can make either 2 large cakes or \\ 35 small cakes/hr. The kitchen is available for 3 hours and 20 large cakes \& 700 small cakes are needed. \\ How many helpers are required? 
 } & \multirow{5}{*}{AQUA} & \multirow{5}{*}{B} & $0.2$ & \textcolor{red}{A ($0.25$)}     \\  
\makecell[cl]{\textbf{Paraphrased}: How many helpers are needed if each helper can make either 2 large cakes or 35 small cakes \\ per hour, and the kitchen is available for 3 hours and needs 20 large cakes and 700 small cakes?\\
\textbf{Options}:\texttt{{[}A)8,B)10,C)12,D)15,E)19{]}}}&  &  & $0.8$ & \textcolor{blue}{B}      \\ \hline

\makecell[cl]{\textbf{Original}: A starts a business with Rs.40,000. After 2 months, B joined him with Rs.60,000. C joined them after \\ some more time with Rs.120,000. At the end of the year, out of a total profit of Rs.375,000, C gets Rs.150,000 \\ as his share. How many months after B joined the business, did C join? 
 } & \multirow{5}{*}{AQUA} & \multirow{5}{*}{B} & $0.1$ & \textcolor{red}{C ($0.4$)}     \\  
\makecell[cl]{\textbf{Paraphrased}: A starts a business with Rs.40,000 and after 2 months, B joins with Rs.60,000. C joins the business \\ at some point later with Rs.120,000. At the end of the year, the total profit is Rs.375,000, and C receives \\ Rs.150,000 as their share. How many months after B joined the business did C join? \\
\textbf{Options}:\texttt{{[}A)2,B)4,C)23,D)24,E)84{]}}} &  &  & $0.45$ & \textcolor{blue}{B}      \\ \hline

\makecell[cl]{\textbf{Original}: A star-polygon is drawn on a clock face by drawing a chord from each number to the fifth number \\ counted clockwise from that number. That is, chords are drawn from 12 to 5, from 5 to 10, from 10 to 3, \\ and so on, ending back at 12. What is the degree measure of the angle at each vertex in the star-polygon?} & \multirow{4}{*}{MATH} & \multirow{4}{*}{"30"} & $0.05$ & \textcolor{red}{"150" ($0.4$)}     \\  
\makecell[cl]{\textbf{Paraphrased}: What is the measure of the angle at each vertex in the star-polygon formed by drawing a \\chord from each number on the clock face to the fifth number counted clockwise from that number?} &  &  & $0.5$ & \textcolor{blue}{"30"}      \\ \hline

\makecell[cl]{\textbf{Original}: By partial fractions, $\frac{1}{ax^2 + bx + c} = \frac{A}{x - \dfrac{-b + \sqrt{b^2 - 4ac}}{2a}} + \frac{B}{x - \dfrac{-b - \sqrt{b^2 - 4ac}}{2a}} $Find $A + B.$} & \multirow{4}{*}{MATH} & \multirow{4}{*}{"0"} & $0.2$ & \textcolor{red}{"1" ($0.25$)}     \\  
\makecell[cl]{\textbf{Paraphrased}: Find the sum of $A$ and $B$ in the expression\\$\frac{1}{ax^2 + bx + c} = \frac{A}{x - \frac{-b + \sqrt{b^2 - 4ac}}{2a}} + \frac{B}{x - \frac{-b - \sqrt{b^2 - 4ac}}{2a}}.$  *Note the \LaTeX code was paraphrased from \texttt{\textbackslash frac} to \texttt{\textbackslash dfrac.}} &  &  & $0.65$ & \textcolor{blue}{"0"}      \\ \hline

\makecell[cl]{\textbf{Original}: Statement 1 | For every positive integer n there is a cyclic group of order n. \\ Statement 2 | Every finite cyclic group contains an element of every order that divides the order of the group. } & \multirow{4}{*}{MMLU} & \multirow{4}{*}{A} & $0.05$ & \textcolor{red}{C ($0.95$)}     \\  
\makecell[cl]{\textbf{Paraphrased}: Statement 1 says that there exists a cyclic group of any positive integer n.\\Statement 2 says that in any finite cyclic group, there is an element for every possible order that divides \\ the order of the group.\\ \textbf{Options}:\texttt{{[}A)True,True, B)False,False, C)True,False, D)False,True{]}}} &  &  & $0.5$ & \textcolor{blue}{A}      \\ \hline

\makecell[cl]{\textbf{Original}: What is the probability that a randomly selected integer in the set $\{1,2,3,\ldots,100\}$ is divisible by 2 \\ and not divisible by 3? Express your answer as a common fraction.} & \multirow{5}{*}{MMLU} & \multirow{5}{*}{D} & $0.25$ & \textcolor{red}{A ($0.4$)}     \\  
\makecell[cl]{\textbf{Paraphrased}: What is the chance that if we randomly choose an integer from the set of numbers 1 to 100, \\ it will be divisible by 2 but not divisible by 3? Write your answer as a fraction.\\ \textbf{Options}:\texttt{{[}A) $\frac{31}{66}$, B) $\frac{17}{66}$, C) $\frac{17}{31}$, D) $\frac{17}{50}${]}}} &  &  & $0.8$ & \textcolor{blue}{D}      \\ \hline

\end{tabular}
\end{adjustbox}
\caption{Qualitative examples where the original problems and corresponding surface form variations exhibit substantial solve rate difference using GPT-3.5-turbo.}
\label{tab:qa1}
\end{table*}

\section{Related Work}
\paragraph{Mathematical Reasoning in LLMs}

The complexity of mathematics necessitates System-2 reasoning, characterized by a slow, step-by-step cognitive process~\cite{system-2}. Numerous works have sought to emulate this process in solving mathematics with LLMs~\cite{cot,SC, cot0, verify-step-by-step, reasoning-surey-many-other-frameworks-still-multi-step}. As a prominent framework, chain-of-thought~\cite{cot,cot0} prompts the language model to generate a sequence of reasoning steps instead of a direct answer; \citet{SC} extended chain-of-thought by Self-Consistency, in which they replaced greedy decoding with sampling decoding to generate a variety of \emph{reasoning paths}, with multiple paths potentially leading to the same answer from different angles. Other multi-step reasoning variations with verifiers exist~\cite{external-reasoning1, got, tot}; however, they are less related to our focus primarily on the language model's internal ability to solve mathematical problems.

\paragraph{Paraphrasing Variability} Previous research on the impact of paraphrasing mathematical problems on their solvability by language learning models (LLMs) is limited. The study by ~\cite{paraphrase-prompt} explored how paraphrased instructions affect the performance of traditional NLP benchmarks. This sensitivity of instructive prompts has inspired further research in prompting learning~\cite{pmp-sensitive, pmp-sensitive2, pmp-sensitive3} and in-context exemplar mechanisms~\cite{ice1,ice2,ice3-unrealiable}. However, our work focuses on the sensitivity of the mathematical problem presentation itself instead of the instruction or in-context examples.

\section{Conclusions}

This work highlights the variability in the solve rate of large-scale language models to the surface form of mathematical problems. Leveraging this, we introduced the Self-Consistency-over-Paraphrases (SCoP), which improves mathematical reasoning performance over Self-Consistency. We hope our findings will inspire the need for more robust language models that can reason effectively regardless of how a problem is presented.

\section*{Limitations}

While we derive thorough conclusions about the relationship between the surface form of a mathematical problem and its solvability by large-scale language models with the effectiveness of SCoP and additional studies, one limitation is the need for a mechanism for identifying or generating surface forms that are easier to solve than others. The study is solely conducted in English, while the generalizability of SCoP in other languages is unexplored. Future research could address these by exploring the rationalization of surface forms, \emph{i.e.}, determining the optimal form given the original one, and the verifiability of the framework in other languages. 

\section*{Ethics Statement}
The datasets that we used in experiments are publicly available. In our work, we explore the relationship between the surface form of a mathematical problem and its solvability by large-scale language models. We do not expect any direct ethical concern from our work.

\section*{Acknowledgements}
This study was supported by MIT-IBM Watson AI Lab.

\bibliographystyle{acl_natbib}
\DeclareRobustCommand{\disambiguate}[3]{#3}
\bibliography{anthology,custom}
\clearpage

\appendix

\section{Choose Margin}\label{sec:app-margin}

We examine the effect of margin on selecting exemplars for in-context paraphrasing using GPT-3.5 and separate dev-sets from GMS8K and MMLU, each with 250 data points. The results in Table~\ref{tab:margin} show that a moderate margin outperforms a large one in SCOP, as the latter may decrease the diversity of exemplars.

\begin{table}[ht!]
\centering
\begin{adjustbox}{width=\columnwidth}

\begin{tabular}{lll}
\hline\hline
Margin & MMLU (Dev, k = 8) & GSM8K (Dev, k = 8) \\ \hline
SC/HPR\% & 53.2 (26.9) / 64 & 73.6 (21.4) / 33.6 \\ \hline
0.2    & 55.6 (34.4)     & 75.2 (31.0)      \\
0.3    & 56.8 (35.6)     & 74.8 (32.1)      \\
0.4    & 55.2 (33.8)     & 75.6 (33.3)      \\
0.5    & 53.6 (32.5)     & 74.4 (35.7)     \\ \hline
\end{tabular}

\end{adjustbox}
\caption{Ablation on the margin effect of exemplar selection.} 
\label{tab:margin}
\end{table}

\section{APE Alternatives}\label{sec:ape}

A potential alternative to finding an optimal prompt for paraphrasing is to use the Automatic Prompt Engineering (APE) settings~\cite{pmp-sensitive2}. We formulate the procedure into four steps:

\begin{enumerate}[]
\setlength\itemsep{0.1em}
\item Present a set of input-output pairs where the inputs are the original problems and the outputs are the paraphrased exemplars. Prompt the language model to generate $C$ candidate instructions that could produce the outputs from the inputs.

\item Prompt each candidate instruction to the language model to generate paraphrases for a batch size $B$ of problems in the development set and compare the mean solve rate change before and after paraphrasing. 

\item Choose the instruction that maximizes the mean solve rate change. 

\item Repeat steps 1 - 3 $E$ times. 

\end{enumerate}

We implemented this procedure using GPT-3.5 on the AQUA development set to obtain the instruction ($C = 15$, $B = 30$, $B = 0$). We tested the performance in both AQUA (in-domain) and GSM8K (out-of-domain), comparing it with ICL$_\text{para}$. Although the in-domain AQUA performance was similar to ICL$_\text{para}$, the out-of-domain performance worsened, and APE required more data than ICL$_\text{para}$. Therefore, this approach has yielded negative results. The performance results are presented in Table~\ref{tab:ape}.

\begin{table}[t!]
\centering
\begin{adjustbox}{width=\columnwidth}

\begin{tabular}{lcccc}
\hline\hline
    & \multicolumn{2}{c}{GSM8K}         & \multicolumn{2}{c}{AQUA}          \\ \hline
    &  ICL$_\text{para}$        & APE             & ICL$_\text{para}$         & APE             \\\hline
SC  & \multicolumn{2}{c}{76.3 (24.5)} & \multicolumn{2}{c}{66.9 (22.2)} \\\hline
N/1 & 77.9 (39.0)   & 73.7 (34.0)   & 66.4 (29.8)   & 66.1 (29.0)   \\
N/2 & 80.5 (39.2)    & 76.3 (36.2)   & 68.5 (31.7)   & 66.8 (30.0)   \\
N/4 & 79.2 (38.3)   & 77.7 (33.0)   & 70.5 (35.4)   & 70.8 (32.0)   \\
N/8 & 80.2 (40.6)   & 79.0 (41.5)    & 69.7 (34.4)   & 69.2 (31.0)  \\ \hline
\end{tabular}

\end{adjustbox}
\caption{A comparison between the performance of APE and ICL$_\text{para}$ paraphrasing.} 
\label{tab:ape}
\end{table}

\section{Temperature and Randomness}

To further validate that SCoP goes beyond simply increasing randomness, we introduce two variants of SC, where we increase the temperature up to 0.9 and 1 using GPT-3.5 on 1/4 of the MMLU and AQuA datasets. The results are shown in Table~\ref{tab:randT}.

\begin{table}[ht!]
\centering
\begin{adjustbox}{width=0.7\columnwidth}

\begin{tabular}{lll}\hline\hline
Temperature & MMLU & AQuA \\\hline
0.7 (baseline)        & 49 (21.5)  & 68 (27.3)        \\
0.9         & 49 (24.6)  & 68 (27.3)        \\
1           & 49 (24.6)  & 64 (18.2)\\ \hline
\end{tabular}

\end{adjustbox}
\caption{Increasing randomness by temperature saturates reasoning performance. The numbers inside the parenthesis are the accuracy for the hard problems.} 
\label{tab:randT}
\end{table}

As can be observed, although increasing temperature brings a slight improvement in the hard problems, the performance gain soon saturates and is not nearly comparable to that of SCoP.

\section{Surface Forms with Solve Rate Degradation}

As previously discussed, surface form modification by paraphrasing can lead to degradation in the solve rate. Here, we present additional qualitative examples where the solution rate worsened after paraphrasing. See Table~\ref{tab:qa-negative}.

\begin{table*}[h]
\begin{adjustbox}{width=\textwidth}

\begin{tabular}{lllll}
\hline
\textbf{Problem} & \textbf{Source}    & \textbf{Label}    & \textbf{SR}  & \textbf{Voted (F.)} \\ \hline
\makecell[cl]{\textbf{Original}: Howie wants to buy cupcakes for everyone in his class as a special treat. He's not sure if \\ people will want vanilla or chocolate cupcakes so he decides to get one of each for everyone. \\If he gets the same amount of 2 cupcakes for each himself, his teacher, and his 25 classmates, \\ how many cupcakes should Howie buy?} & \multirow{4}{*}{GSM8K} & \multirow{4}{*}{"54"} & $0.8$ &   -   \\  
\makecell[cl]{\textbf{Paraphrased}: Howie wants to purchase cupcakes for his entire class as a special treat. \\ Since he is unsure of the flavor preference, he plans to buy both vanilla and chocolate cupcakes. \\ Howie wants to ensure that he has an equal amount of cupcakes for himself, his teacher, and \\ his 25 classmates. How many cupcakes should Howie purchase in total? } &  &  & $0.25$ & \textcolor{red}{"27" (0.35)}      \\ \hline

\makecell[cl]{\textbf{Original}: Janice bikes at 10 miles per hour, while Jennie bikes at 20. How long until they have \\ collectively biked 1 mile?
 } & \multirow{5}{*}{AQUA} & \multirow{5}{*}{B} & $0.55$ & -    \\  
\makecell[cl]{\textbf{Paraphrased}: Janice and Jennie are biking at different speeds. Janice bikes at a rate of 10 miles per hour, \\ while Jennie bikes at a rate of 20 miles per hour. How much time will it take for them to collectively \\ bike a distance of 1 mile? \\
\textbf{Options}:\texttt{{[}A)1 minute, B)2 minutes, C)3 minutes, D)4 minutes, E)5 minutes{]}}}&  &  & $0.1$ & \textcolor{red}{C (0.35)}     \\ \hline

\makecell[cl]{\textbf{Original}: How many primes are in the row of Pascal's Triangle that starts with a $1$ followed by a $6$?} & \multirow{3}{*}{MATH} & \multirow{3}{*}{"0"} & $0.7$ & -  \\  
\makecell[cl]{\textbf{Paraphrased}: Starting with the numbers $1$ and $6$, how many prime numbers are there  in the \\  sequence of numbers in Pascal's Triangle?} &  &  & $0.1$ & \textcolor{red}{"2" (0.25)}      \\ \hline

\makecell[cl]{\textbf{Original}: John divided his souvenir hat pins into two piles. The two piles had an equal number of pins.\\  He gave his brother one-half of one-third of one pile. John had 66 pins left. How many pins did John \\ originally have?} & \multirow{4}{*}{MMLU} & \multirow{4}{*}{B} & $0.9$ & -     \\  
\makecell[cl]{\textbf{Paraphrased}: John started with a certain number of souvenir hat pins. He divided them into two equal piles. \\He then gave his brother one-half of one-third of one of the piles. After that, John was left with 66 pins. \\How many pins did John have at the beginning?\\ \textbf{Options}:\texttt{{[}A) 396, B) 72, C) 66, D) 36{]}}} &  &  & $0.35$ & \textcolor{red}{A (0.4)}      \\ \hline

\end{tabular}
\end{adjustbox}
\caption{Qualitative examples where paraphrased surface forms of the original problems can also exhibit solve rate degradation, using GPT-3.5-turbo. }
\label{tab:qa-negative}
\end{table*}

\newpage

\section{Prompt Templates}\label{sec:temp}

We list the prompt templates used in the paper below. 
\begin{tcolorbox}[
    colback=white,     
    colframe=gray!75,
    title=Na\"ive Paraphrasing
]
Paraphrase the following math problem: \{\texttt{target problem}\}
\end{tcolorbox}

\begin{tcolorbox}[
    colback=white,     
    colframe=gray!75,
    title=ICL Paraphrasing
]
Paraphrase the following math problem: \{\texttt{input problem}\} \\
Output: \{\texttt{Paraphrased exemplar}\} \\
(Repeat $N_{shot}$)
\tcblower
Paraphrase the following math problem: \{\texttt{target problem}\}
\end{tcolorbox}

\begin{tcolorbox}[
    colback=white,     
    colframe=gray!75,
    title=APE Candidate Search
]
A student is completing a task that requires producing a text output from a text input. The student receives instruction about several rules that describe how to produce the outputs given the inputs. What is the instruction?
\end{tcolorbox}

\begin{tcolorbox}[
    colback=white,     
    colframe=gray!75,
    title=Few-shot Chain-of-thought]
    
Question: At Academic Academy, to pass an algebra test you must score at least 80. If there are 35 problems on the test, what is the greatest number you can miss and still pass? \\
Answer Choices: A) 7 B) 28 C) 35 D) 8 \\
Rationale: First, we need to find 80\% of 35. We can do this by multiplying 35 by $0.80$: $35 \times 0.80 = 28$. So, if you get 28 problems correct, you will have scored 80\% on the test.\\
To find the greatest number you can miss and still pass, subtract the number you can get correct from the total number of problems:$35 - 28 = 7$. \\

Therefore, the greatest number you can miss and still pass is (A) 7.  \\
(Repeat $N_{shot}$)
\tcblower
Question: \{\texttt{target problem}\}
\end{tcolorbox}


\end{document}